\documentclass[10pt,twocolumn,letterpaper]{article}

 \usepackage{cvpr}              

\usepackage{amsmath}    
\usepackage{amssymb}    
\usepackage{mathtools}   
\usepackage{MnSymbol}
\usepackage[ruled,vlined]{algorithm2e}
\usepackage{float}
\usepackage{amsmath}
\DeclareMathOperator*{\E}{\mathbb{E}}
\DeclarePairedDelimiter{\norm}{\|}{\|}
\newcommand{\keypoint}[1]{\vspace{0.1cm}\noindent\textbf{#1}\quad}
\definecolor{cvprblue}{rgb}{0.21,0.49,0.74}
\usepackage[pagebackref,breaklinks,colorlinks,allcolors=cvprblue]{hyperref}

\title{Test-Time Alignment of Text-to-Image Diffusion Models \\ via Null-Text Embedding Optimisation}

\author{
    Taehoon Kim$^{1}$\thanks{Correspondence to: Taehoon Kim, Mr.Chow@ed.ac.uk},\quad
    Henry Gouk$^{1}$,\quad
    Timothy Hospedales$^{1,2}$ \\
    \\
    $^{1}$School of Informatics, University of Edinburgh \\
    $^{2}$Samsung AI Center, Cambridge
}

\begin{document}
\maketitle
\begin{abstract}
Test-time alignment~(TTA) aims to adapt models to specific rewards during inference. However, existing methods tend to either under-optimise or over-optimise (reward hack) the target reward function.  We propose \textbf{Null-Text Test-Time Alignment (Null-TTA)}, which aligns diffusion models by optimising the unconditional embedding in classifier-free guidance, rather than manipulating latent or noise variables. 
Due to the  structured semantic nature of the text embedding space, this ensures alignment occurs on a semantically coherent manifold and prevents reward hacking (exploiting non-semantic noise patterns to improve the reward).
Since the unconditional embedding in classifier-free guidance serves as the anchor for the model's generative distribution, Null-TTA directly steers model's generative distribution towards the target reward rather than just adjusting the samples, even without updating model parameters. 
Thanks to these desirable properties, we show that Null-TTA achieves state-of-the-art target test-time alignment while maintaining strong cross-reward generalisation. This establishes semantic-space optimisation as an effective and principled novel paradigm for TTA.
\end{abstract}

\section{Introduction}

Diffusion models~\cite{ho2020ddpm,song2021score,rombach2022ldm} have demonstrated remarkable ability in modelling complex data distributions across various domains, including images~\cite{dhariwal2021beatgans,saharia2022photorealistic} and language~\cite{li2022diffusion}. Despite their generative power, these models are trained on large-scale, uncurated web datasets that often contain undesirable or misaligned content. Consequently, aligning pre-trained diffusion models with human values or target objectives—such as aesthetic quality or preference—is essential for trustworthy deployment of diffusion models in real world applications.

Existing approaches for alignment can be broadly divided into two categories: \textit{fine-tuning based alignment} and \textit{test-time alignment (TTA)}.  
(1) Fine-tuning methods~\cite{black2024training, zhang2024confronting, lee2023aligning, prabhudesai2023aligning, wallace2024diffusion, fan2023dpok, clark2023directly, dong2023raft} directly modify model parameters to optimise a reward function but are computationally expensive and prone to \textit{reward over-optimisation}, where the model overfits to proxy rewards and loses generalisation across multiple rewards or output diversity. (2) TTA methods optimise latent or noise variables to maximise a differentiable reward functions~\cite{chungdiffusion,kim2025das,xie2025dymo,tang2025dno,guo2024initno, he2024manifold,li2025dynamic,ramesh2025test}; sample from intractable reward-conditioned posteriors via techniques such as Sequential Monte Carlo~(SMC)~\cite{kim2025das, dou2024diffusion, wu2023practical, singhal25general}; or explore denoising trajectories using discrete search algorithms~\cite{li2025dynamic,ramesh2025test, ma2025inference}. But these suffer from reward under/over-optimisation due to highly unstructured nature of latent/noise spaces, inefficiency of SMC and vast search space. 

To address these limitations, 
we draw inspiration from recent advances in diffusion model editing that leverage optimisation in the text-conditioning space. In particular, \textit{Null-Text Inversion (NTI)}~\cite{mokady2023null} demonstrated that optimising the unconditional, or \emph{null-text}, embedding in Classifier-Free Guidance (CFG)~\cite{ho2022classifier} allows fine-grained semantic control while preserving image fidelity.  
Building on this insight, we extend the principle of null-text optimisation beyond image editing and reformulate it as a general mechanism for reward alignment in diffusion models.

To be specific, we propose \textbf{Null-Text Test-Time Alignment (Null-TTA)}, a novel TTA framework for text-to-image diffusion models that effectively aligns pre-trained models without suffering from either under- or over-optimisation. 
Instead of manipulating latent or noise variables, we optimise the \textit{unconditional (null) text embedding} within CFG, which serves as the geometric anchor for the conditional generative distribution.
By shifting optimisation to this structured semantic space, Null-TTA ensures optimisation occurs on a structured  semantically manifold. This \emph{prevents the reward-hacking phenomenon} -- generating non-semantic noise patterns that optimise the target reward only -- while degrading other reward metrics, sample diversity, and even subjective visual quality. 
Null-TTA combined with our principled objective realigns the model’s generative distribution towards the target reward conditioned distribution, rather than simply correcting the samples in a way that increases target reward, without updating model parameters. This ensures effective and efficient alignment to target rewards.

Our main contributions are summarised as follows:
\begin{itemize}
    \item \textbf{Null-Text Test-Time Alignment (Null-TTA)}, a training-free framework that performs alignment in the structured semantic space of text conditioning.
    \item A principled objective that reorients the model's generative distribution itself via null-text embedding optimisation, enabling stable and efficient reward alignment without parameter updates.
    \item Extensive experiments demonstrate that Null-TTA achieves state-of-the-art target reward performance and strong generalisation across diverse reward functions.
\end{itemize}

\section{Related Work}

\subsection{Test-Time Alignment (TTA)}
TTA methods align diffusion model outputs with desired rewards during inference, without further training.  
They can be grouped into three main branches: guidance-based, sampling-based, and search-based approaches.

\keypoint{Guidance-based methods.}  
These methods~\cite{tang2025dno,guo2024initno,he2024manifold, chungdiffusion} optimise latent or noise representations to increase differentiable reward scores.  
Direct Noise Optimisation (DNO)~\cite{tang2025dno} manipulates injected noise for reward maximisation, and InitNO~\cite{guo2024initno} tunes initial noise vectors for semantic consistency.  
DPS~\cite{chungdiffusion} approximates reward posterior sampling by guiding samples at each time step using reward gradients and Manifold Preserving Guided Diffusion (MPGD)~\cite{he2024manifold} constrains guidance within data manifolds to maintain coherence.  
However, optimisation in unstructured latent or noise spaces can still incur off the data manifold behaviour, producing unstable and semantically irrelevant updates that may lead to reward hacking and visually unappealing samples.  
\textbf{Null-TTA} addresses this issue by performing optimisation within the structured semantic space of the text encoder, which provides implicit manifold regularisation and enables stable, reward-consistent alignment.

\keypoint{Sampling-based methods.}  
Sampling based methods~\cite{kim2025das, wu2023practical, dou2024diffusion, singhal25general} aim to sample from a reward-aligned posterior using techniques such as Sequential Monte Carlo~(SMC).  
While effective, these methods implicitly assume a fixed posterior distribution, which is intractable,  and focus on sampling efficiency rather than direct distributional optimisation. Also, sampling itself can be inaccurate when using a small number of particles — a common choice for scalability.
\textbf{Our Null-TTA} departs from this paradigm by explicitly modifying the generative distribution itself, offering stronger and more consistent reward alignment across tasks.

\keypoint{Search-based methods.}  
Recent works such as DSearch~\cite{li2025dynamic}, Search over Paths~\cite{ma2025inference} and Noise Trajectory Search~\cite{ramesh2025test} cast TTA as a discrete searching problem in noise space, employing search algorithms such as Monte Carlo Tree Search (MCTS).  
Although they can handle non-differentiable rewards, their search complexity scales exponentially with diffusion steps and dimension of latents. This leads to slow inference speed and sub-optimal performances. 
\textbf{Null-TTA} instead performs continuous optimisation within the conditioning manifold, combining efficiency with smooth, interpretable control over reward objectives.

\subsection{Fine-Tuning Based Alignment}
Fine-tuning methods~\cite{black2024training, zhang2024confronting, lee2023aligning, prabhudesai2023aligning, wallace2024diffusion, fan2023dpok, clark2023directly, dong2023raft} adapt diffusion models towards human preference or task-specific rewards by updating model weights.
RL-based methods~\cite{lee2023aligning, fan2023dpok, black2024training, dong2023raft} adapt reward or policy optimisation to diffusion models for improved alignment, while gradient-based approaches~\cite{clark2023directly, prabhudesai2023aligning, wallace2024diffusion} directly backpropagate reward signals for stable and efficient tuning.
TDPO~\cite{zhang2024confronting} further addresses reward overoptimisation to enhance robustness.
Although these methods achieve strong reward alignment, they require extensive computation and risk \textit{reward over-optimisation}, degrading diversity.  
Consequently, recent research—including ours—focuses on TTA frameworks that maintain generality and efficiency without costly fine-tuning.

\section{Methods}
\subsection{Preliminaries}
\subsubsection{Diffusion Models}
Diffusion models~\cite{ho2020ddpm,song2021score} learn to model complex data distributions through a forward–reverse stochastic process.  
In the forward process, Gaussian noise is gradually added to a clean sample $x_0$ over $T$ steps according to
\begin{equation}
    q(x_t|x_{t-1}) = \mathcal{N}\left( \sqrt{1-\beta_t}x_{t-1}, \beta_t \mathbf{I} \right),
\end{equation}
where $\beta_t$ controls the noise schedule.  
This yields a closed form $q(x_t|x_0) = \mathcal{N}(\sqrt{\bar{\alpha}_t}x_0, (1-\bar{\alpha}_t)\mathbf{I})$ with $\bar{\alpha}_t = \prod_{i=1}^t\alpha_t=\prod_{i=1}^t (1-\beta_i)$.

The reverse process is learned via a neural network $\epsilon_\theta(x_t, t, c)$ that predicts the noise added at step $t$, conditioned on textual or visual input $c$.  
The denoising process then follows
\begin{equation}
    p_\theta(x_{t-1}|x_t, c) = \mathcal{N}(\tilde{\mu}_t(x_t, c), \sigma_t^2 \mathbf{I}),
\end{equation}
where $\tilde{\mu}_t(x_t, c)$ is derived from the predicted clean sample $\hat{x}_0$ as:
\begin{equation}
\tilde{\mu}_t(x_t,c)
=
\frac{\sqrt{\bar{\alpha}_{t-1}}\beta_t}{1-\bar{\alpha}_t}\hat{x}_0
+
\frac{\sqrt{\alpha_t}(1-\bar{\alpha}_{t-1})}{1-\bar{\alpha}_t}x_t ,
\end{equation}

where $\hat{x}_0$ is defined as: 
\begin{equation}
    \hat{x}_0 = \frac{1}{\sqrt{\bar{\alpha}_t}}\left(x_t - \sqrt{1-\bar{\alpha}_t}\epsilon_\theta(x_t, t, c)\right).
\end{equation}

\subsubsection{Latent Diffusion Models (LDMs).}
Latent Diffusion Models~(LDMs)~\cite{rombach2022ldm} extend DDPMs by operating within the latent space of a pre-trained autoencoder, enabling efficient yet high-fidelity generation.  
Specifically, an encoder $\mathcal{E}(\cdot)$ maps an image $x \in \mathcal{X}$ to a latent code $z = \mathcal{E}(x)$, and a decoder $\mathcal{D}(\cdot)$ reconstructs the input as $\mathcal{D}(\mathcal{E}(x)) \approx x$.  
The diffusion model is then trained on the latent space $\mathcal{Z}$ rather than pixel space, producing denoised latents $z_0$ that can be decoded back to image domain.

The denoising process operates on $z_t$ instead of $x_t$, and can be conditioned on additional modalities such as text.  
In the case of Stable Diffusion~\cite{rombach2022ldm}, the condition is a text embedding $c = f_{\text{CLIP}}(y)$ produced by a pre-trained CLIP text encoder~\cite{radford2021learning}.  
The training objective of the LDM is then expressed as a simplified noise prediction loss:
\begin{equation}
    \mathcal{L} = 
    \E_{z \sim \mathcal{E}(x),\, c,\, \epsilon \sim \mathcal{N}(0,I),\, t}
    \left[ \| \epsilon - \epsilon_\theta(z_t, c, t) \|_2^2 \right],
\end{equation}
where $z_t = \sqrt{\bar{\alpha}_t}z + \sqrt{1-\bar{\alpha}_t}\epsilon$.  
During inference, the latent $z_T \sim \mathcal{N}(0,1)$ is iteratively denoised to obtain $z_0$, which is then decoded as $\mathcal{D}(z_0)$ to synthesise the final image.

\subsubsection{Classifier-Free Guidance (CFG)}
Classifier-Free Guidance~\cite{ho2022classifier} enhances conditional generation by extrapolating between unconditional and conditional predictions:
\begin{equation}
    \tilde{\epsilon}_\theta(x_t, t, c, \phi) = \epsilon_\theta(x_t, t, \phi) 
    + s \big(\epsilon_\theta(x_t, t, c) - \epsilon_\theta(x_t, t, \phi)\big),
\end{equation}
where $s$ is a guidance scale and $\phi$ denotes the \emph{null-text} embedding used for unconditional generation.  
The null embedding $\phi$ anchors the model’s generative distribution in a structured semantic space.  

\subsubsection{Tweedie’s Formula}
For a Gaussian denoising process, the posterior mean of the clean sample given $x_t$ can be expressed using Tweedie’s formula~\cite{chungdiffusion}:
\begin{equation}
    \E[x_0|x_t, \phi] = \frac{1}{\sqrt{\bar{\alpha}_t}}\left(x_t - \sqrt{1-\bar{\alpha}_t}\,\E[\epsilon|x_t, \phi]\right),
\end{equation}
where $\E[\epsilon|x_t, \phi]$ is approximated by the model’s noise prediction $\tilde{\epsilon}_\theta(x_t,t,c, \phi)$.  
This formula provides an efficient way to estimate $\hat{x}_0(x_t, \phi) = \E[x_0|x_t, \phi]$ during sampling and is central to our optimisation framework.

\subsection{Alignment of Diffusion Models}
Aligning diffusion models with a reward function $r(x)$ can be viewed as finding a new target distribution $p_{\text{tar}}$ that maximises the expected reward:
\begin{equation}
    p_{\text{tar}} = \arg\max_p \E_{x\sim p}[r(x)].
\end{equation}
However, this direct optimisation can cause \textit{reward hacking}, leading to distributional drift from the pretrained model~\cite{gao2023scaling}.  
To prevent this, KL regularisation is introduced~\cite{uehara2024fine}:
\begin{equation}
\label{eq:tar}
    p_{\text{tar}} = \arg\max_p \E_{x\sim p}[r(x)] - \alpha D_{\mathrm{KL}}(p||p_{\text{pre}}),
\end{equation}
where $p_{\text{pre}}$ is the pretrained diffusion model distribution.  
Following~\cite{rafailov2024dpo}, the optimal target distribution takes the form
\begin{equation}
    p_{\text{tar}}(x) = \frac{1}{\mathcal{Z}} p_{\text{pre}}(x)\exp\left(\frac{r(x)}{\alpha}\right),
\end{equation}
where $\mathcal{Z}$ is a normalisation constant.  

Most TTA methods can be seen as trying to \emph{sample} \cite{kim2025das,wu2023practical, dou2024diffusion, singhal25general}, or else \emph{maximise} \cite{tang2025dno,li2025dynamic, xie2025dymo, ramesh2025test, he2024manifold, chungdiffusion, guo2024initno} $p_{\text{tar}}(x)$. Importantly, both families of approaches sample or maximise in either pixel-space $x$ or, almost equivalently, latent-space $z$ -- both of which are high-dimensional unstructured spaces. This makes them prone to relying on non-semantic noise perturbations to satisfy $r(x)$. 

\subsection{Null-Text Test-Time Alignment (Null-TTA)}
\label{sec:ntta}
We argue that optimising the null-text embedding instead of the noise itself provides a more structured and efficient way of manipulating existing diffusion models, ultimately resulting in better test-time alignment. Unlike unstructured latent or noise variables, the null-text embedding resides in a semantically meaningful space defined by the text encoder. This structured embedding space imposes an implicit manifold regularisation, ensuring that the optimisation moves in coherent semantic directions, unlike latent or noise-space optimisation that can exploit noisy non-semantic perturbations in $x$ to maximise the reward.

Following the idea from Eq.~(\ref{eq:tar}), we design a regularised objective that steers the model's generative distribution towards reward-maximising direction within the semantic manifold of the original model.
To be specific, we regularise the reward maximisation by penalising distributional deviation between the data distribution induced by the optimised null-text embedding and that of the original model. This encourages the model to remain close to its pretrained behaviour while still achieving meaningful reward alignment. The resulting objective is formulated as:
\begin{equation}
\begin{split}
\max_{\phi'} \Big( \lambda_1 \E_{p(x_0|\phi')} [R(x_0)] & - \lambda_2 \operatorname{KL}(p(x_{0:T},\phi')||p(x_{0:T},\phi))\Big)
\end{split}
\end{equation}
where $\phi$ and $\phi'$ denote the original and optimised null-text embeddings respectively, and $R(\cdot)$ is a target reward function. For simplicity, we omit time step $t$ and conditioning text $c$.

Using the Markov property of the diffusion process, the above can be decomposed into a sum of local KL divergences between consecutive denoising steps, and KL divergence between embedding distributions. Intuitively, these terms constrain the model to preserve the consistency of its denoising trajectory with respect to the pretrained generative process while preventing the optimised embedding moving too far from the original null text embedding. After simplification, the objective can be written as:
\begin{align}
& \operatorname{KL}(p(x_{0:T},\phi')||p(x_{0:T}, \phi)) \\
&= \operatorname{KL}\Bigg( p(\phi')p(x_T|\phi')\prod_{i=1}^{T} p(x_{i-1}|x_i, \phi') \nonumber \\
& \mspace{36mu} || p(\phi)p(x_T|\phi)\prod_{i=1}^{T} p(x_{i-1}|x_i, \phi) \Bigg) \\
&= \operatorname{KL}\Bigg( p(\phi')p(x_T)\prod_{i=1}^{T} p(x_{i-1}|x_i, \phi') \nonumber \\
& \mspace{36mu} || p(\phi)p(x_T)\prod_{i=1}^{T} p(x_{i-1}|x_i, \phi) \Bigg) \\
&= \E_{p(x_{0:T},\phi')} \Bigg[ \log \frac{p(\phi')\prod_{i=1}^{T} p(x_{i-1}|x_i, \phi')}{p(\phi)\prod_{i=1}^{T} p(x_{i-1}|x_i, \phi)} \Bigg] \\
&= \E_{p(x_{0:T},\phi')} \Bigg[\log \frac{p(\phi')}{p(\phi)}+ \sum_{i=1}^{T}\log \frac{p(x_{i-1}|x_i, \phi')}{ p(x_{i-1}|x_i, \phi)} \Bigg] \\
&= \sum_{i=1}^{T} \E_{p(x_i|\phi')} \left[ \operatorname{KL}(p(x_{i-1}|x_i, \phi')||p(x_{i-1}|x_i, \phi)) \right] \\
&\quad \quad \quad \quad \quad+ \operatorname{KL}(p(\phi')||p(\phi)).
\end{align}

Each conditional distribution in the diffusion process is Gaussian, parameterised by the denoising prediction network. Therefore, we can write:
\begin{align}
p(x_{i-1}|x_i, \phi') &= \mathcal{N}\Big( \tilde{\mu}_i(x_i,\phi'), \sigma_i^2 \mathbf{I} \Big),
\end{align}
where
\begin{equation}
\tilde{\mu}_i(x_i,\phi') \coloneqq \frac{\sqrt{\bar{\alpha}_{i-1}}\beta_i}{1-\bar{\alpha}_i} \hat{x}_0(x_i,\phi') + \frac{\sqrt{\alpha_i}(1-\bar{\alpha}_{i-1})}{1-\bar{\alpha}_i} x_i,
\end{equation}
\begin{equation}
\hat{x}_0(x_i,\phi') = \frac{1}{\sqrt{\bar{\alpha}_i}}\left(x_i - \sqrt{1-\bar{\alpha}_i}\tilde{\epsilon}(x_i, \phi')\right).
\end{equation}

This leads to the simplified closed-form of the KL term:
\begin{align}
\operatorname{KL}&(p(x_{i-1}|x_i, \phi')||p(x_{i-1}|x_i, \phi)) \\
&= \frac{1-\alpha_i}{2\alpha_i(1-\bar{\alpha}_i)} \norm{\tilde{\epsilon}(x_i, \phi')-\tilde{\epsilon}(x_i, \phi)}^2.
\end{align}

Furthermore, we model $\phi$ and $\phi'$ as Gaussian random variables $\mathcal{N}(\phi, \sigma_{\phi}^2I_d)$ and $\mathcal{N}(\phi', \sigma_{\phi}^2I_d)$, respectively, leading to
\begin{equation}
\operatorname{KL}(p(\phi')||p(\phi))=\frac{1}{2\sigma_\phi^2}\norm{\phi-\phi'}^2 .
\end{equation}
Note that this probabilistical modeling is for intuitive interpretation and we do not sample null text embedding but just optimise mean of this distribution~($\phi'$) for computational efficiency.

Combining all terms, the final optimisation objective becomes:
\begin{equation} 
\begin{split} 
\max_{\phi'} \Big( &\lambda_1 \E_{p(x_0|\phi')} [R(x_0)] \\& - \lambda_2\sum_{t=1}^{T}\frac{1-\alpha_t}{2\alpha_t(1-\bar{\alpha}_t)} \norm{\tilde{\epsilon}(x_t, \phi')-\tilde{\epsilon}(x_t, \phi)}^2 \\ & - \frac{\lambda_2}{2\sigma_\phi^2} \|\phi'-\phi\|^2 \big) . 
\end{split} 
\end{equation}

In practice, we approximate the expectation via Monte Carlo integration over a single denoising trajectory and apply Tweedie’s formula to estimate $E_{p(x_0|\phi')} [R(x_0)]$. Thus, at each time step $t$, we optimise:
\begin{equation}
\label{eq:main_obj}
\begin{split}
\max_{\phi'} \Big( \lambda_{1} R(\hat{x}_0(x_t,\phi')) & - \frac{\lambda_{2}(1-\alpha_t)}{2\alpha_t(1-\bar{\alpha}_t))}\norm{\tilde{\epsilon}(x_t, \phi')-\tilde{\epsilon}(x_t, \phi)}^2 \\
& - \frac{\lambda_2}{2\sigma_\phi^2} \norm{\phi'-\phi}^2 \Big).
\end{split}
\end{equation}
Empirically, we find that strong regularisation in early noise steps stabilises optimisation, whereas weaker regularisation in later denoising steps enables finer alignment. Therefore, $\lambda_2$ is adaptively annealed as $t \rightarrow 0$ to ensure smooth convergence toward the reward aligned sample distribution. Similarly, the number of optimisation steps increases from $n_\text{min}$ to $n_\text{max}$ with denoising steps since the reward approximation  becomes  more accurate and reliable as denoising proceeds.

\subsection{Greedy Search for the Backward Process}

At each denoising step $t$, we augment the standard DDPM update with a lightweight
\textit{greedy search} mechanism that deterministically selects the
most reward-promising latent for the next timestep.  After updating the
unconditional embedding $\phi'$ at timestep $t$, we draw $K$ candidates
$\{x_{t-1}^{(k)}\}_{k=1}^K \sim p(x_{t-1}\mid x_t,\phi')$ from the DDPM
transition kernel.  For each candidate, we estimate its corresponding clean
sample using Tweedie's posterior mean,
\[
    \hat{x}_0^{(k)} = \frac{1}{\sqrt{\bar\alpha_{t-1}}}
    \left(x_{t-1}^{(k)} - \sqrt{1-\bar\alpha_{t-1}}\,\tilde\epsilon(x_{t-1}^{k}, \phi')\right),
\]
and score it with the reward model $R(\hat{x}_0^{(k)})$.
We then keep only the highest-scoring candidate as $x_{t-1}$ and discard the
rest. This greedy selection nudges the reverse diffusion path toward
higher-reward regions by exploring $K$ candidates at each step and keeping the
best one, yielding a more reward-aligned trajectory.
Detailed procedure for the proposed method is illustrated in the Supplementary materials.

\begin{figure*}[t]
    \centering
    \includegraphics[width=\textwidth]{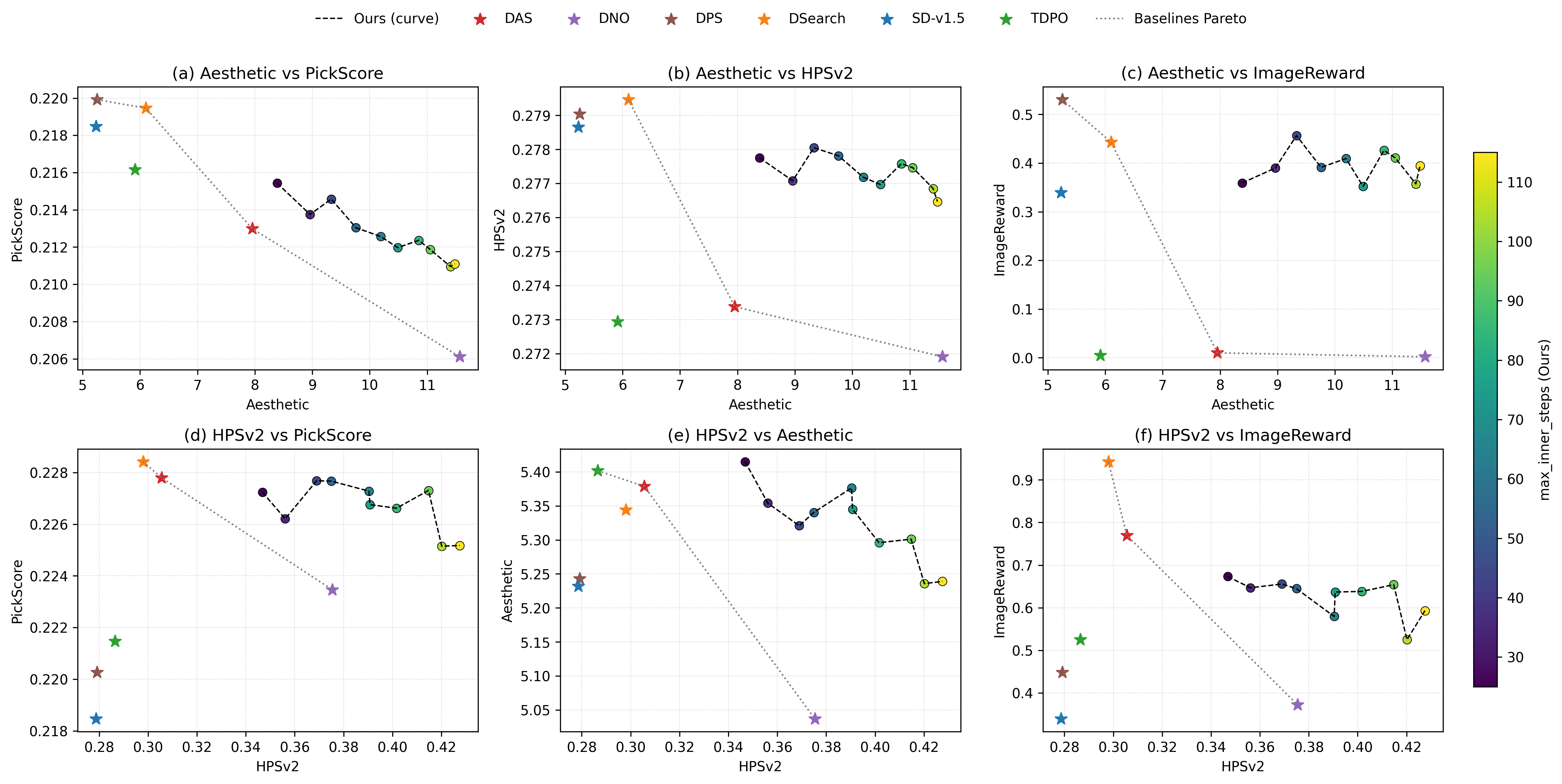} 
    \caption{Evaluation of reward-optimisation (x-axes) vs over-optimisation (generalisation  to held-out rewards on y-axes). Top/Bottom rows correspond to aligning with Aesthetic and HPSv2 target rewards respectively.  
    Each $\filledstar$ point corresponds to a particular baseline method, which together define the state-of-the-art pareto front (dashed line). For Null-TTA, each $\circ$ point indicates the optimisation intensity (maximum inner steps $n_\text{max}$).  Null-TTA consistently shifts the Pareto front outward, indicating superior target reward optimisation while maintaining strong generalisation across unseen rewards.}
    \label{fig:pareto}
    \vspace{-2mm}
\end{figure*}

\begin{table}[t]
\centering
\caption{
Target reward optimisation versus cross-reward generalisation for PickScore target. Null-TTA ($n_{\text{max}}=55$) outperforms competitors in both target and held-out rewards.
}
\setlength{\tabcolsep}{5pt} 
\renewcommand{\arraystretch}{1.0} 
\scalebox{0.85}{
\begin{tabular}{@{}lcccc@{}}
\toprule
Method   & PickScore & HPSv2 & Aesthetic & ImageReward\\
\midrule
\multicolumn{5}{l}{\textbf{Target: PickScore}}\\ \midrule
SD-v1.5~\cite{rombach2022ldm}  & 0.218 & 0.279 & 5.232  & 0.339 \\
DNO~\cite{tang2025dno}        & 0.289 & 0.290 & 5.075 & 0.396 \\
DAS~\cite{kim2025das}          & 0.258 & 0.289 & 5.382 & 0.871  \\
Null-TTA                        & \textbf{0.315} & \textbf{0.294} & 
\textbf{5.431} & \textbf{0.946} \\
\bottomrule
\end{tabular}
}\vspace{-2mm}
 \label{tab:main}
\end{table}

\begin{figure*}[t]
    \centering
    \includegraphics[width=0.9\linewidth]{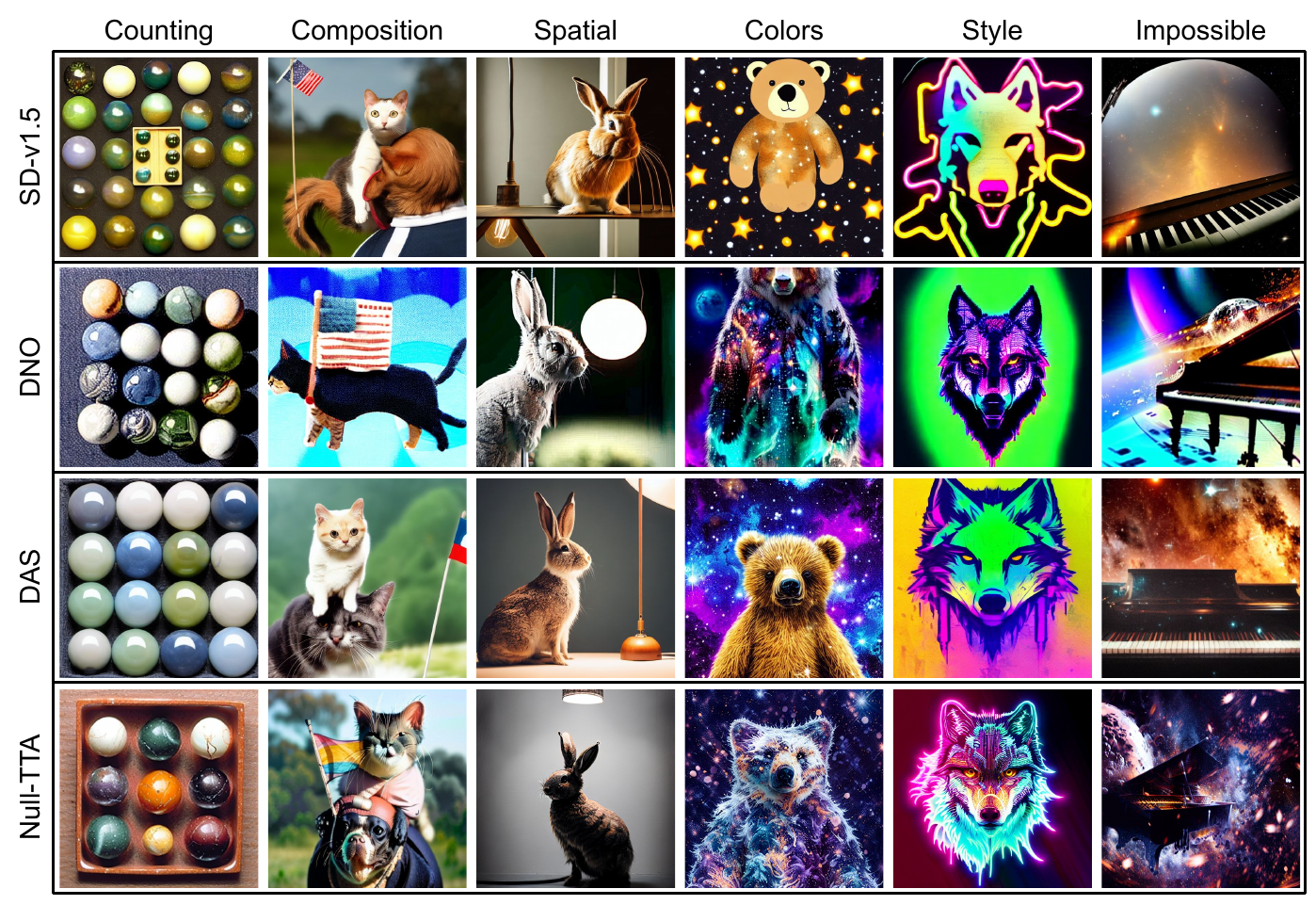}
    \caption{
Qualitative comparison on six challenging categories—counting
(\textit{“Nine marbles arranged in a perfect square”}), compositionality
(\textit{“A cat riding on a dog's back while holding a tiny flag”}),
spatial reasoning (\textit{“A rabbit standing directly under a hanging lamp”}),
unusual colors (\textit{“A bear with galaxy-patterned fur”}),
fine-grained style transfer (\textit{“A wolf drawn in vibrant cyberpunk neon edge highlights”}),
and impossible scenes (\textit{“A piano drifting in deep space surrounded by comets”}).
Null-TTA consistently produces images that more faithfully satisfy the
prompt constraints while preserving global coherence.
}
    \label{fig:qualitative}
\end{figure*}

\section{Experiments}
\subsection{Experimental Setup}
\keypoint{Metrics}
Throughout the experiments, we evaluate (1)~Improvements on the target reward used for optimisation and (2)~Cross-reward generalisation ability. For target rewards, we employ HPSv2~\cite{wu2023human}, PickScore~\cite{kirstain2023pick} and Aesthetic score~\cite{LAIONAestheticPredictorV1}. For evaluation, we use HPSv2~\cite{wu2023human}, Aesthetic score~\cite{LAIONAestheticPredictorV1}, Pickscore~\cite{kirstain2023pick} and ImageReward~\cite{xu2023imagereward}. We adapted Stable Diffusion (SD) v1.5~\cite{rombach2022ldm} as our baseline model. All the experiments were done with the prompts from HPDv2 datasets~\cite{wu2023human} unless specified otherwise. We used a single NVIDIA L40S GPU for all experiments.
Note that all the results are averaged over 3 different random seeds. Implementation details including hyperparameters can be found at supplementary materials.

\keypoint{Comparison Methods}
We choose competitor methods to cover all the existing diffusion alignment branches: TDPO~\cite{zhang2024confronting} for Finetuning based; DNO~\cite{tang2025dno} and DPS~\cite{chungdiffusion} for guidance based TTA; DAS~\cite{kim2025das} for sampling based TTA; and DSearch~\cite{li2025dynamic} for search based TTA. All the methods were reproduced using official implementation and their hyper-parameter settings except that the inference steps are set to 100.

\subsection{Alignment Without Over-Optimisation} We first evaluate the ability of Null-TTA and competitors to align diffusion samples with a target reward, while avoiding over-optimisation as measured by degradation on other held-out rewards. We follow \cite{kim2025das} evaluation approach by analysing the Pareto trade-offs between each target metric (HPSv2~\cite{wu2023human} or Aesthetic score~\cite{LAIONAestheticPredictorV1}) and the remaining evaluation scores (PickScore~\cite{kirstain2023pick} and ImageReward~\cite{xu2023imagereward}). As shown in Fig.~\ref{fig:pareto}: (1) \emph{Null-TTA achieves similar or better target reward than all competitors} (Right-most point Null-TTA's curve), while offering  better generalisation to held out rewards than the best competitor, DNO~\cite{tang2025dno}. This indicates successful alignment without over-optimisation. (2) \emph{Null-TTA pareto-dominates competitor methods} Null-TTA's $n_\text{max}$ hyperparameter defines how aggressively it optimises the target reward and thus enables a range of trade-offs between target and held out rewards. However, it's noteworthy that increased Null-TTA target reward optimisation only weakly degrades held out rewards, enabling Null-TTA's pareto-domination of competitors. This also reflects successful alignment without over-optimisation. Note that $n_{\text{min}}$ was set to 5 and $n_{\text{max}}$ was set to $\{25, 35, 45, 65, 75, 85, 95, 105, 115\}$.


We attribute these results to existing TTA baselines either exhibiting reward over-optimisation -- improving target metric while degrading others (eg: DNO~\cite{tang2025dno}), or under-optimisation -- weak improvement in the target metric (eg: TDPO~\cite{zhang2024confronting}, DPS~\cite{chungdiffusion}). In contrast, Null-TTA maintains performance across both optimised and held out objectives due to (i) Optimisation on the manifold of the null-text embedding, which naturally constrains updates to meaningful directions and prevents the model from exploiting non-semantic artefacts; and (ii) our KL-regularised objective explicitly preserves local consistency of the denoising trajectory with respect to the pretrained model, preventing excessive distributional drift during alignment.


We also extend the above analysis to consider PickScore~\cite{kirstain2023pick} as the seen reward for optimisation. The results in Table~\ref{tab:main} show that Null-TTA achieves superior target reward optimisation as well as held out reward generalisation. 


\begin{figure}[t] 
    \centering
    \includegraphics[width=0.9\columnwidth]{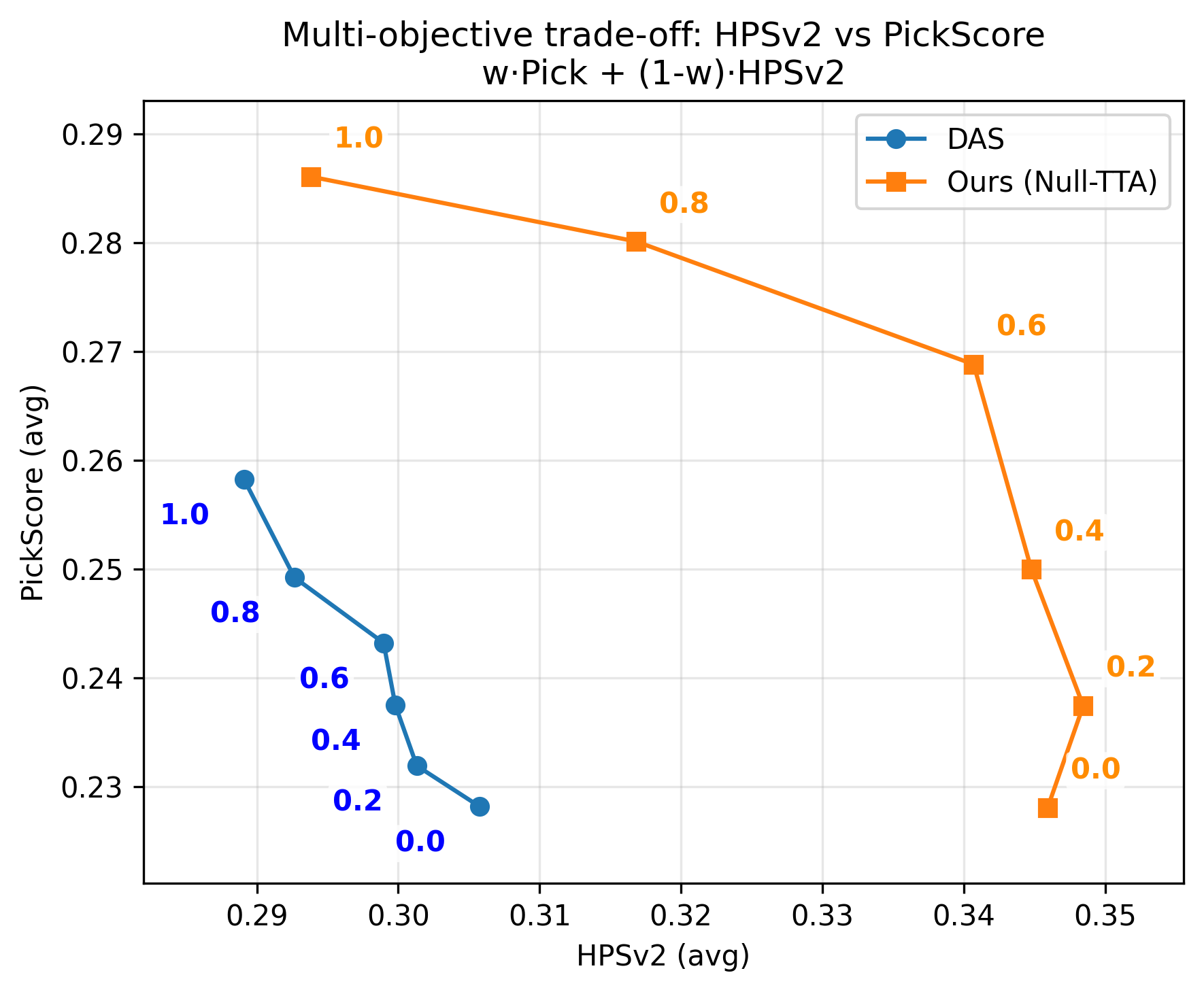}
    \vspace{-2mm}
    \caption{
Multi-objective optimisation using 
\(R_{\text{multi}} = w \cdot \text{PickScore} + (1-w)\cdot \text{HPSv2}\). 
Each marker corresponds to a different weight \( w \).
}
\vspace{-2mm}
    \label{fig:multi-objective-tradeoff}
\end{figure}

\begin{table}[t!]
\centering
\caption{User study on perceptual quality (16 users, 800 pairs).}
\setlength{\tabcolsep}{10pt} 
\renewcommand{\arraystretch}{1.0} 
\scalebox{0.85}{ 
\begin{tabular}{lcccc}
\toprule
\textbf{Method} & \textbf{1st (Best)} & \textbf{2nd} & \textbf{3rd} & \textbf{Rank $\downarrow$} \\ 
\midrule
DAS~\cite{kim2025das} & 26.9\% & 31.0\% & 42.1\% & 2.15 \\
DNO~\cite{tang2025dno} & 30.9\% & 34.4\% & 34.7\% & 2.04 \\ 
\midrule
\textbf{Ours} & \textbf{42.2\%} & \textbf{34.6\%} & \textbf{23.2\%} & \textbf{1.81} \\ 
\bottomrule
\end{tabular}
}\vspace{-2mm} 

\label{tab:user_study}
\end{table}

\subsection{Qualitative Results}
We qualitatively evaluate Null-TTA on six categories that are known to be particularly challenging for
diffusion models—counting, compositionality, spatial reasoning, unusual colors,
fine-grained style transfer, and impossible scenes~\cite{kim2025das}. Across all categories, baseline models exhibit
typical failure modes such as incorrect object counts, entangled relationships, spatial
misplacement, or collapsing the intended semantics into a more plausible interpretation. 

In contrast, Null-TTA consistently generates images that closely adhere to the prompt-specific
constraints while preserving global coherence. These improvements align with our overall findings:
optimising in the semantic null-text space with our principled objective function yields
reward-aligned updates that avoid over-optimisation and faithfully capture fine-grained prompt
requirements. Representative examples are shown in Figure~\ref{fig:qualitative}. All photos were generated $(n_{\text{min}}, n_{\text{max}}) = (5, 25)$ and PickScore~\cite{kirstain2023pick} as target reward.
We also conducted a blind user study on HPDv2~\cite{wu2023human} (Table~\ref{tab:user_study}, 16 participants, 800 evaluations). Humans statistically prefer Null-TTA over baselines, confirming our reward gains translate to perceptual quality.

\subsection{Multi-Objective Optimisation}
We further evaluate whether Null-TTA can effectively balance multiple competing objectives by 
optimising a weighted combination of HPSv2~\cite{wu2023human} and PickScore~\cite{kirstain2023pick},
\[
R_{\text{multi}} = w \cdot \text{PickScore} + (1-w)\cdot \text{HPSv2}.
\]

Figure~\ref{fig:multi-objective-tradeoff} compares the resulting trade-off curves of Null-TTA and DAS~\cite{kim2025das}. 
Both methods exhibit smooth frontiers as the weight \( w \) varies, but pareto frontier of Null-TTA clearly dominates that of the baseline.  
These results highlight that semantic-space optimisation enables Null-TTA to perform substantially better multi-objective alignment than existing TTA approaches.

\subsection{Computational Cost}
Table~\ref{tab:cost} compares the computational cost of Null-TTA to baseline TTA methods under the HPSv2 target~\cite{wu2023human}. 
Across all settings, Null-TTA achieves stronger target-reward performance than competing methods with comparable inference-time budgets. Notably, it runs faster than DNO~\cite{tang2025dno}—the strongest baseline in terms of reward optimisation—while using the least GPU memory among all methods.
This efficiency arises from our design choice to update only the null-text embedding, which requires backpropagation solely through the cross-attention layers rather than the entire U-Net. 
By avoiding full-network gradient computation, Null-TTA maintains low memory usage and stable runtime scaling as the number of optimisation steps increases.

\begin{figure*}[t!] 
    \centering
    \includegraphics[width=\linewidth]{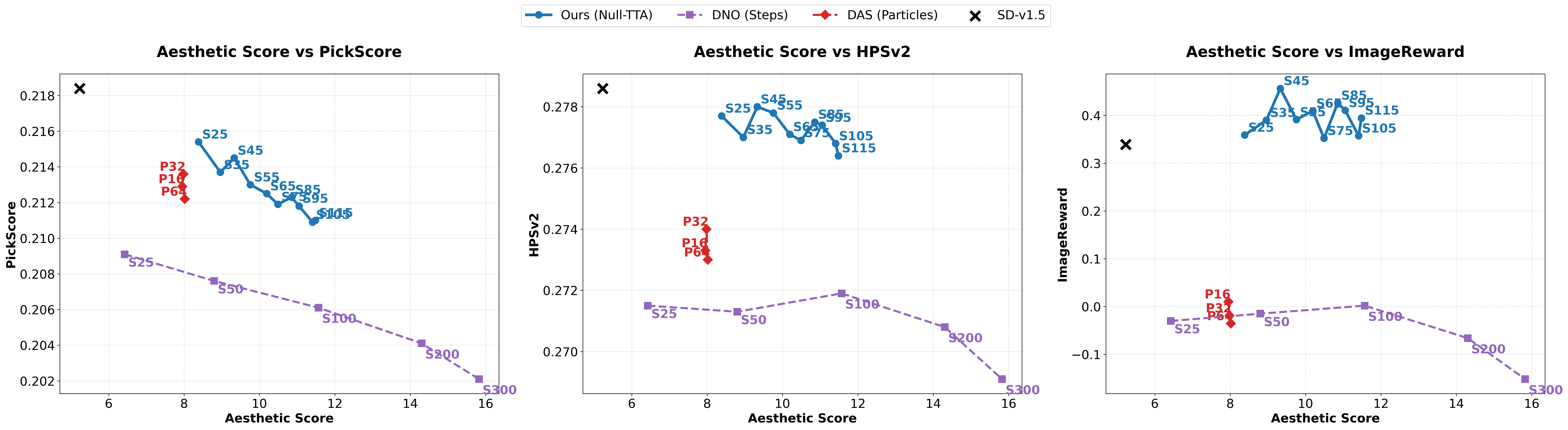}
    \vspace{-6mm}
    \caption{Comparison across various computational budgets. P/S refers to number of particles/optimization steps.}
    \label{fig:pareto_aesthetic}
    \vspace{-2mm}
\end{figure*}

Although Null-TTA outperforms existing baselines even with lower GPU memory and wall-clock time usage, we conducted experiments with different computational budgets for DAS~\cite{kim2025das} and DNO~\cite{tang2025dno}.
As shown in Figure~\ref{fig:pareto_aesthetic}, Null-TTA clearly shows better pareto curve compared to the baselines. 
Note that default settings for DAS~\cite{kim2025das} and DNO~\cite{tang2025dno} are P16 and S100 respectively.

\begin{table}[t]
\centering
\caption{
Computational comparison of Null-TTA and baseline TTA methods under the HPSv2 target. 
For each method, we report GPU memory usage, wall-clock time per generated image, and the resulting HPSv2 score. 
Null-TTA achieves stronger target-reward performance than baselines under similar inference budgets, 
while also exhibiting the lowest memory consumption. 
}
\begin{scalebox}{0.85}{
\begin{tabular}{lcccc}
\toprule
\textbf{Method} 
& $\mathbf{n_{\text{max}}}$  
& \textbf{HPSv2}
& \textbf{GPU (MB)} 
& \textbf{Time} \\
\midrule
DAS~\cite{kim2025das} 
& -- 
& 0.306
& 30595 
& 4m48s \\
DNO~\cite{tang2025dno} 
& -- 
& 0.375
& 20449 
& 19m38s \\
\midrule
Null-TTA (Ours) 
& 25 
& 0.347
& \textbf{17585} 
& \textbf{4m33s} \\
Null-TTA (Ours) 
& 55 
& 0.375
& \textbf{17585}
& 8m40s \\
Null-TTA (Ours) 
& 85 
& 0.402
& \textbf{17585}
& 12m44s \\
Null-TTA (Ours) 
& 115
& \textbf{0.428}
& \textbf{17585}
& 17m07s \\
\bottomrule
\end{tabular}
}
\end{scalebox}

\label{tab:cost}
\vspace{-2mm}
\end{table}

\subsection{Results with SDXL}
To further validate the effectiveness of Null-TTA, we provide results with large-scale diffusion models, specifically SDXL~\cite{podell2023sdxl}. We utilized the official implementations for DAS~\cite{kim2025das} and DNO~\cite{tang2025dno}. 

As shown in Table~\ref{tab:pickscore_comparison}, Null-TTA consistently outperforms baselines in the target reward (PickScore~\cite{kirstain2023pick}) while maintaining high scores in cross-reward metrics (HPSv2~\cite{wu2023human}, Aesthetic~\cite{LAIONAestheticPredictorV1}, ImageReward~\cite{xu2023imagereward}). Notably, as the optimization intensity ($n_{max}$) increases from 25 to 45, our method achieves higher target rewards without significant degradation in other quality metrics, demonstrating robust alignment capabilities.

\begin{table}[t]
    \centering
        \caption{Target reward optimization versus cross-reward generalization for PickScore target on SDXL. We compare our method with different optimization steps ($n_{max}$) against baselines. All scores are averaged over 3 random seeds.}
        \vspace{-2mm}
    \label{tab:pickscore_comparison}
    \resizebox{\linewidth}{!}{%
    \begin{tabular}{lcccc}
        \toprule
        Method & PickScore $\uparrow$ & HPSv2 $\uparrow$ & Aesthetic $\uparrow$ & ImageReward $\uparrow$ \\
        \midrule
        \multicolumn{5}{l}{\textbf{Target: PickScore}} \\
        \midrule
        SDXL~\cite{podell2023sdxl}& 0.218 & 0.279 & 5.236 & 0.338 \\
        DNO~\cite{tang2025dno}       & 0.222 & 0.278 & 5.231 & 0.057 \\
        DAS~\cite{kim2025das}        & 0.253 & 0.290 & 5.545 & 1.260 \\
        \midrule
        Null-TTA ($n_{max}=25$) & 0.266 & 0.291 & \textbf{5.622} & 1.224 \\
        Null-TTA ($n_{max}=45$) & \textbf{0.282} & \textbf{0.293} & 5.613 & \textbf{1.276} \\
        \bottomrule
    \end{tabular}
    }
\vspace{-2mm}
\end{table}

\subsection{Results with Non-differentiable Reward Functions}
Many real-world applications, especially in scientific domains, rely on simulation-based evaluations where the gradient of the objective function is inaccessible or computationally prohibitive to obtain. Examples include evaluating compressibility algorithms (e.g., JPEG size), molecular docking scores in drug discovery, or fluid dynamics simulations. In these scenarios, the reward function $r(\mathbf{x})$ behaves as a black box. We demonstrate that the proposed method can effectively align the diffusion model outputs with such non-differentiable rewards.

To overcome the non-differentiability, we employ a zeroth-order gradient estimation strategy. Specifically, let $\phi$ denote the optimization variable (e.g., the null-text embedding). We estimate the gradient $\nabla_{\phi} J(\phi)$ as below:

\begin{equation}
    \hat{\nabla}_{\phi} J(\phi) \approx \frac{1}{K \mu} \sum_{k=1}^{K} \left[ J(\phi + \mu \mathbf{v}_k) - J(\phi) \right] \mathbf{v}_k
\end{equation}

where $J(\cdot)$ is an objective function, $K$ is the number of random perturbation samples, $\mu$ is the perturbation scale, and $\mathbf{v}_k \sim \mathcal{N}(\mathbf{0}, \mathbf{I})$ are random Gaussian vectors.

For the JPEG compressibility experiment, we set the hyperparameters as follows: $\lambda_1 = 50$, $\lambda_2 = 0.002$, and $\sigma_\phi^2 = 0.01$. For the zeroth-order gradient estimation, we used a perturbation scale of $\mu=0.02$ and a sample count of $K=4$.

We applied this method to optimize the JPEG compressibility (file size) as a proxy for non-differentiable scientific objectives. As shown in Table \ref{tab:jpeg_comparison}, Null-TTA achieves a competitive compression rate compared to DNO~\cite{tang2025dno}, while significantly preserving the visual quality and prompt alignment as evidenced by higher scores in PickScore~\cite{kirstain2023pick}, HPSv2~\cite{wu2023human}, Aesthetic~\cite{LAIONAestheticPredictorV1} and Imagereward~\cite{xu2023imagereward} metrics. Note that we do not utilise any gradient information while DNO~\cite{tang2025dno} utilises gradient of diffusion model.

\begin{table}[t]
    \centering
    \caption{Quantitative comparison on the non-differentiable \textbf{JPEG Compressibility} target. We report the JPEG Reward (defined as negative file size in kB) as the target metric, alongside cross-reward generalization metrics. Higher JPEG Reward indicates smaller file size. Null-TTA ($n_{max}=25$) is compared against baselines. All scores are averaged over 3 random seeds.}
    \label{tab:jpeg_comparison}
    \resizebox{\linewidth}{!}{%
    \begin{tabular}{lccccc}
        \toprule
        Method & JPEG Reward & PickScore  & HPSv2  & Aesthetic  & ImageReward  \\
        \midrule
        \multicolumn{6}{l}{\textbf{Target: JPEG Compressibility}} \\
        \midrule
        SD-v1.5~\cite{rombach2022ldm}   & -81.496 & 0.218 & 0.279 & 5.236 & 0.338 \\ \midrule
        DNO~\cite{tang2025dno}          & \textbf{-30.132}  & 0.190 & 0.236 & 3.988 & -1.541 \\
        
        Null-TTA         & -33.741  & \textbf{0.204} & \textbf{0.253} & \textbf{4.649} & \textbf{-0.893} \\
        \bottomrule
    \end{tabular}
    }
    \vspace{-4mm}
\end{table}

\section{Conclusion}
We present Null-Text Test-Time Alignment (Null-TTA), a training-free framework that aligns text-to-image diffusion models by optimising the null-text embedding in classifier-free guidance. By shifting optimisation from unstructured latent or noise spaces to the semantic manifold of the text encoder, Null-TTA provides inherent regularisation that prevents reward hacking and maintains semantic coherence.

We also derive a principled objective that directly adjusts the model’s generative distribution towards reward while preserving the pretrained model's knowledge. With this principled objective, our method achieves state-of-the-art target reward optimisation, strong cross-reward generalisation, and favorable computational efficiency across diverse settings.

Our findings show that optimising within the semantic conditioning space offers a simple yet powerful alternative to existing TTA approaches, and establishes Null-TTA as a principled paradigm for reliable inference-time alignment of diffusion models.

\paragraph{Acknowledgements}
Taehoon Kim was supported by the Engineering and Physical Sciences Research Council and Ministry of Defence Centre for Doctoral Training in Sensing, Processing and AI for Defence and Security, [EP/Y013859/1]. This project was supported by the Royal Academy of Engineering under the Research Fellowship programme.
{
    \small
    \bibliographystyle{ieeenat_fullname}
    \bibliography{main.bib}
}

\clearpage
\appendix
\twocolumn[
\begin{center}
    {\LARGE \textbf{Test-Time Alignment of Text-to-Image Diffusion Models}\\[0.5em]
     \LARGE \textbf{via Null-Text Embedding Optimisation} \par}
    \vspace{1.0em}
    {\large Supplementary Material \par}
\end{center}
\vspace{1.5em}
]

\begin{table}[t!]
\centering
\caption{Influence of Search Width on Null-TTA rewards.} 
\vspace{-2mm}
\begin{scalebox}{0.85}{
\begin{tabular}{c|cccc}
\toprule
\textbf{Search Width} 
& \textbf{PickScore} 
& \textbf{Aesthetic} 
& \textbf{HPSv2} 
& \textbf{ImageReward} \\
\midrule
1   & 0.226 & 5.340 & 0.339 & 0.585 \\
3   & 0.228 & 5.449 & 0.346 & \textbf{0.741} \\
10  & 0.228 & \textbf{5.453} & 0.347 & 0.684 \\
25  & \textbf{0.229} & 5.335 & \textbf{0.352} & 0.700 \\
50  & 0.228 & 5.370 & 0.351 & 0.671 \\
\bottomrule
\end{tabular}
}
\end{scalebox}

\label{tab:particle}
\vspace{-2mm}
\end{table}

\begin{table}[t!]
\centering
\caption{Comparison of optimised scores for different values of annealing coefficient ($\gamma$).} 
\vspace{-2mm}
\begin{scalebox}{0.85}{
\begin{tabular}{c|cccc}
\toprule
\textbf{$\gamma$} 
& \textbf{PickScore} 
& \textbf{AestheticScore} 
& \textbf{HPSv2} 
& \textbf{ImageReward}\\
\midrule
0.004 & 0.225  & 5.404 & 0.318 & 0.664 \\
0.008 & \textbf{0.228}  & \textbf{5.449} & 0.346 & \textbf{0.741} \\
0.016 & 0.226  & 5.266 & \textbf{0.399} & 0.700 \\
\bottomrule
\end{tabular}
}
\end{scalebox}
 \label{tab:gamma}
\vspace{-4mm}
\end{table}

\begin{algorithm*}[t!]
\caption{Null-Text Test-Time Alignment with Greedy Search (Null-TTA)}
\label{alg:null_tta_exact}
\SetKwInOut{Input}{Input}
\SetKwInOut{Output}{Output}
\SetKw{KwTo}{to}

\Input{Pre-trained diffusion model $\epsilon_\theta$, Reward Function $R(\cdot)$, Prompt $\mathcal{P}$, Total steps $T$, Step bounds $n_{min}, n_{max}$, Growth rate $\gamma$, Weights $\lambda_{1}, \lambda_{2}, \sigma_{\phi}^2$, Search Width $K$, Learning rate $\eta$.}
\Output{Aligned image $x_0$}

Initialize latent $x_T \sim \mathcal{N}(0, \mathbf{I})$\;
Get initial embeddings: $\phi \leftarrow \text{TextEncoder}(\emptyset)$, $c \leftarrow \text{TextEncoder}(\mathcal{P})$\;
Initialize optimizable embedding: $\phi' \leftarrow \phi$\;

\For{$t = T, \dots, 1$}{
    \tcp{1. Dynamic Hyperparameter Scheduling}
    Calculate adaptive regularization weight: $\tilde{\lambda}_{2,t} \leftarrow \lambda_2 \cdot \max(2 - (1+\gamma)^{T-t}, 0)$\;
    Calculate adaptive optimization steps: $N_t \leftarrow n_{min} + \lfloor \min((1+\gamma)^{T-t}-1, 1) \cdot (n_{max} - n_{min}) \rfloor$\;
    
    \tcp{2. Null-Text Optimization (Eq. 26)}
    \For{$n = 1, \dots, N_t$}{
        Compute unconditional noise: $\epsilon_{uncond} \leftarrow \epsilon_\theta(x_t, t, \phi')$\;
        Compute original unconditional noise (fixed): $\epsilon_{orig} \leftarrow \epsilon_\theta(x_t, t, \phi)$\;
        Compute original CFG noise: $\tilde{\epsilon}_{orig} \leftarrow \epsilon_{orig} + s(\epsilon_\theta(x_t, t, c) - \epsilon_{orig})$\;
        Compute CFG noise: $\tilde{\epsilon} \leftarrow \epsilon_{uncond} + s(\epsilon_\theta(x_t, t, c) - \epsilon_{uncond})$\;
        
        Estimate clean image (Tweedie's Formula): $\hat{x}_0(x_t, \phi') \leftarrow \frac{x_t - \sqrt{1-\bar{\alpha}_t}\tilde{\epsilon}}{\sqrt{\bar{\alpha}_t}}$\;
        
        Calculate Objective $J(\phi')$:
        \begin{equation*}
            J(\phi') = \lambda_{1}R(\hat{x}_{0}(x_{t},\phi^{\prime})) - \frac{\tilde{\lambda}_{2,t}(1-\alpha_{t})}{2\alpha_{t}(1-\overline{\alpha}_{t})}||\tilde{\epsilon}-\tilde{\epsilon}_{orig}||^{2} -\frac{\tilde{\lambda}_{2,t}}{2\sigma_{\phi}^{2}}||\phi^{\prime}-\phi||^{2}
        \end{equation*}
        
        Compute Gradient: $g \leftarrow \nabla_{\phi'} J(\phi')$ (via Autograd or Zeroth-Order Estimation)\;
        Update Embedding: $\phi' \leftarrow \phi' + \eta \cdot g$ \tcp*{Gradient Ascent}
    }
    
    \tcp{3. Greedy Search}
    Generate $K$ candidates $\{x_{t-1}^{(k)}\}_{k=1}^K$ from $p_\theta(x_{t-1}|x_t, \phi')$\;
    Evaluate candidates via Tweedie's estimate: $r^{(k)} \leftarrow R(\hat{x}_0(x_{t-1}^{(k)}))$ for $k=1 \dots K$\;
    Select best trajectory: $x_{t-1} \leftarrow x_{t-1}^{(k^*)}$ where $k^* = \arg\max_k r^{(k)}$\;
}
\Return{$x_{final} = \text{Decoder}(x_0)$}\;
\end{algorithm*}

\section{Implementation Details} For Null-TTA, we set $(\lambda_1, \lambda_2, \sigma_{\phi}^2)$ as $(100, 0.002, 0.01)$ for HPSv2~\cite{wu2023human}, PickScore~\cite{kirstain2023pick} and $(2, 0.002, 0.01)$ for Aesthetic score~\cite{LAIONAestheticPredictorV1}. As stated in Sec~3.3, we gradually decrease $\lambda_2$ as denoising proceeds with annealing coeficient $\max(2-(1+\gamma)^{T-t}, 0)$. Similarly, as reward approximation by Tweeidie's formula becomes accurate as denoising proceeds, we increase the number of the optimisation steps gradually. optimisation step at each iteration is computed as $n_{\text{opt}}=n_{\text{min}}+\lambda_{\text{opt}}^t*(n_{\text{max}}-n_{\text{min}})$ where $\lambda_{\text{opt}}^t=\min((1+\gamma)^{T-t}-1, 1)$. We set $n_\text{min}$ to 5 and conduct experiments with varying $n_\text{max}$. $\gamma$ is set to 0.008 and exploration number for greedy search $K$ is set to 3. We used Adam optimiser~\cite{kingma2014adam} with learning rate 0.01 for optimisation. Note that all the results are averaged results from 3 different random seeds.

\section{Ablations}
We ablate the key components of Null-TTA to understand how optimisation stability and 
reward–generalisation trade-offs are influenced by the search width, annealing schedule, 
and the balance between the reward and regularisation terms. 
Across all experiments (Tables~\ref{tab:particle},~\ref{tab:gamma},~\ref{tab:regbalance}), 
Null-TTA consistently exhibits stable behaviour: target rewards improve while cross-reward 
generalisation is largely preserved. 
Each hyperparameter controls a different aspect of the optimisation dynamics, and together they 
illustrate why Null-TTA avoids the reward over- and under-optimisation commonly observed in prior TTA methods. All the experiments were done with $(n_{\text{min}}, n_{\text{max}}) = (5, 25)$ and HPSv2 as target reward.

\subsection{Effect of Search Width.}
Table~\ref{tab:particle} reports the effect of search width used in the 
latent-space greedy search step. 
The search width does not influence the optimisation of the null-text embedding itself; 
instead, it controls the extent of latent-space search after optimisation. 

A small search width (e.g., 3) provides a mild but useful search, 
allowing the sampler to more reliably follow the optimisation-improved update direction. 
This leads to modest improvements in both the target reward (HPSv2) and the cross rewards. Excessively large search width broadens the search too aggressively. This can pull samples away from the semantic manifold and reintroduce 
undesired latent variations, eroding cross-reward metrics even as the target reward continues to improve -- an effect reminiscent of mild over-optimisation. Notably, even without search, Null-TTA already produces strong performance, showing that Null-TTA does not depend on greedy search to achieve alignment. 
We set search width to 3 as the default, as it balances search with stability.

\subsection{Effect of the annealing coefficient $\gamma$.}
Table~\ref{tab:gamma} evaluates the role of the annealing coefficient~$\gamma$. 
Importantly, $\gamma$ influences both (i) the rate at which the optimisation strength increases, 
and (ii) the decay speed of the regularisation terms in Eq.~26. 

When $\gamma$ is too large, the optimisation steps grow rapidly while the regularisation decays quickly, 
causing the model to emphasise the (potentially inaccurate) early reward approximation. 
This can lead to early-stage over-optimisation and reduced stability. 
In contrast, very small $\gamma$ values slow both optimisation growth and regularisation decay; it limits the ability to sufficiently 
optimise the target reward even when the reward approximation becomes accurate at later timesteps.
Null-TTA maintains stable behaviour across all tested values of $\gamma$, and moderate settings 
(e.g., $\gamma=0.008$) achieve the best balance between strong HPSv2 improvement and cross-reward retention.

\subsection{Effect of the reward–regularisation balance.}
Table~\ref{tab:regbalance} examines how the balance between the reward term and the two 
regularisation components in Eq.~26 affects optimisation behaviour. 
Increasing the regularisation strength (larger $\lambda_2$ or smaller $\sigma_\phi^2$) leads to 
more conservative updates: target–reward improvement becomes limited, but cross-reward 
generalisation improves, indicating an under-optimised yet stable solution. 
Conversely, weakening the regularisation results in more aggressive updates that favour the 
target reward but can destabilise cross-reward metrics, resembling the early symptoms of 
reward over-optimisation.

Our default configuration achieves a balanced trade-off, providing strong target reward gains 
while maintaining high cross-reward performance. This demonstrates that Null-TTA jointly 
improves alignment and generalisation when the reward–regularisation ratio is within a 
reasonable range, and that Null-TTA remains robust even as this ratio varies.

\begin{table}[t]
\centering
\caption{Effect of the reward/regularisation ratio on the target reward (HPSv2) 
and the cross reward (ImageReward). 
All rows except SD-v1.5 share the same target weight $\lambda_\alpha = 100$.}
\vspace{-2mm}
\scalebox{0.85}{%
\begin{tabular}{lccc}
\toprule
$(\lambda_1, \lambda_2, \sigma_\phi^2)$ 
& HPSv2 
& ImageReward 
& Note \\
\midrule
$(100, 0.0004, 0.01)$  & 0.343 & 0.642 & Weak reg ($\lambda_{\text{reg}}\!\downarrow$) \\
$(100, 0.0100, 0.01)$  & 0.339 & \textbf{0.787} & Strong reg ($\lambda_{\text{reg}}\!\uparrow$) \\
$(100, 0.0020, 0.050)$ & 0.340 & 0.303 & Weak reg ($\sigma_\phi^2\!\uparrow$) \\
$(100, 0.0020, 0.002)$ & 0.337 & 0.781 & Strong reg ($\sigma_\phi^2\!\downarrow$) \\
\midrule
$(100, 0.0020, 0.010)$ & \textbf{0.346} & 0.741 & Ours (default) \\
\midrule
SD-v1.5 & 0.279 & 0.332 & Baseline \\
\bottomrule
\end{tabular}%
} \label{tab:regbalance}
\vspace{-4mm}
\end{table}

\section{Detailed Procedure for Null-TTA}
We provide the detailed algorithmic procedure of our proposed Null-TTA in Algorithm \ref{alg:null_tta_exact}.


\end{document}